\documentclass{article}

\usepackage{arxiv}

\usepackage[utf8]{inputenc} 
\usepackage[T1]{fontenc}    
\usepackage{hyperref}       
\usepackage{url}            
\usepackage{booktabs}       
\usepackage{amsfonts}       
\usepackage{nicefrac}       
\usepackage{microtype}      
\usepackage{lipsum}		
\usepackage{graphicx}
\usepackage{natbib}
\usepackage{doi}
\usepackage{array}
\usepackage{alltt}          
\usepackage{fixltx2e}       
\usepackage{multirow}
\usepackage[table,xcdraw]{xcolor}
\usepackage{diagbox}

\newcolumntype{R}[1]{>{\raggedleft\let\newline\\\arraybackslash\hspace{0pt}}m{#1}}

\title{CLASSLA-Stanza: The Next Step for Linguistic Processing of South Slavic Languages}

\author{Luka Terčon \\
	Faculty of Computer and Information Science\\
	University of Ljubljana\\
	Večna pot 113, 1000 Ljubljana, Slovenia \\
	\texttt{luka.tercon@fri.uni-lj.si} \\
	\And
	Nikola Ljubešić \\
	Jožef Stefan Institute\\
	Jamova cesta 39, 1000 Ljubljana, Slovenia\\
	\texttt{nikola.ljubesic@ijs.si} \\}

\renewcommand{\shorttitle}{CLASSLA-Stanza}

\hypersetup{
pdftitle={A template for the arxiv style},
pdfsubject={q-bio.NC, q-bio.QM},
pdfauthor={David S.~Hippocampus, Elias D.~Striatum},
pdfkeywords={First keyword, Second keyword, More},
hidelinks
}

\begin{document}
\maketitle

\begin{abstract}

We present CLASSLA-Stanza, a pipeline for automatic linguistic annotation of the South Slavic languages, which is based on the Stanza natural language processing pipeline. We describe the main improvements in CLASSLA-Stanza with respect to Stanza, and give a detailed description of the model training process for the latest 2.1 release of the pipeline. We also report performance scores produced by the pipeline for different languages and varieties. CLASSLA-Stanza exhibits consistently high performance across all the supported languages and outperforms or expands its parent pipeline Stanza at all the supported tasks. We also present the pipeline's new functionality enabling efficient processing of web data and the reasons that led to its implementation.

\end{abstract}


\section{Introduction}


The South Slavic languages make up one of the three major branches of the Slavic language family. Despite their widespread use, many members of this group remain relatively low-resourced and under-represented in the field of natural language processing. \cite{corpuscoll} include Macedonian and Bosnian in their list of languages that are significantly under-resourced despite having more than 1 million speakers. 

Although much additional work is required if South Slavic languages are ever to become capable of competing with linguistic giants such as English, steps have already been taken towards establishing common platforms for supporting the development of new resources and tools for these languages. The CLARIN  Knowledge  centre  for  South  Slavic  languages (CLASSLA\footnote{\url{https://www.clarin.si/info/k-centre/}}), was established as a result of prior cooperation in the development of language resources for Slovenian, Croatian, and Serbian, and currently acts as a platform providing expertise and support for developing language resources for South Slavic languages~\citep{classla}. The efforts of the knowledge centre gave rise to the CLASSLA-Stanza\footnote{\url{https://github.com/clarinsi/classla}} pipeline for linguistic processing, which arose as a fork of the Stanza neural pipeline~\citep{qi2020stanza}. CLASSLA-Stanza was created with the aim of providing state-of-the-art automatic linguistic processing for South Slavic languages~\citep{ljubesic-dobrovoljc-2019-neural}, and currently supports  Slovenian, Croatian, Serbian, Bulgarian, and Macedonian. Additionally, Slovenian, Croatian, and Serbian have support for both standard and non-standard varieties. In comparison to its Stanza parent pipeline, CLASSLA-Stanza expands to cover the standard Macedonian language, as well as the non-standard, Internet varieties of Slovenian, Croatian and Serbian. Beside the expanded coverage of languages and varieties, CLASSLA-Stanza shows improvements in performance on all presented levels.

The aim of this paper is to provide both a systematic overview of the differences that CLASSLA-Stanza has to the official Stanza pipeline and a description of the model training procedure which was adopted when training models for the latest 2.1 release. The description of the training procedure is intended to serve as the main reference for future releases as well as for anyone using the CLASSA-Stanza tool to produce their own models for linguistic annotation.

In accordance with this aim, we first describe the differences between CLASSLA-Stanza and Stanza in section \ref{sec:clasvsstan}. Afterwards, section \ref{sec:datasets} introduces the datasets used for training the models. Section \ref{sec:modtrapro} then gives a detailed description of the model training process, which is followed by an analysis of the results produced by the newly-trained models in section \ref{sec:erran}.

At present, the CLASSLA-Stanza annotation tool supports a total of six tasks: tokenization, morphosyntactic annotation, lemmatization, dependency parsing, semantic role labeling, and named-entity recognition. Tokenization is handled by one of two external rule-based tokenizers included in CLASSLA-Stanza, either the Obeliks tokenizer\footnote{\url{https://github.com/clarinsi/obeliks}} for standard Slovenian~\citep{obeliks} or the ReLDI tokenizer\footnote{\url{https://github.com/clarinsi/reldi-tokeniser}} for non-standard Slovenian and all other languages~\citep{reldi-tokenizer}. While the basic tasks of tokenization, morphosyntactic annotation, lemmatization and dependency parsing are covered at least for some languages in the parent Stanza pipeline, semantic role labeling and named entity recognition for South Slavic languages are available only in CLASSLA-Stanza.

 The current version of the models was trained on data that are annotated according to three separate systems for morphosyntactic annotation: the universal part-of-speech tags and the universal morphosyntactic features tags—which are both part of the Universal Dependencies framework for grammatical annotation~\citep{UDtheoretical} and will henceforth be referred to as UPOS and UFeats—and the MULTEXT-East V6 specifications for morphosyntactic annotation~\citep{multext-east}, which are implemented as the language-specific XPOS tags in the CoNLL-U file format\footnote{\url{https://universaldependencies.org/format.html}}, the central file format used by CLASSLA-Stanza.  For dependency parsing, the Universal Dependencies system for syntactic dependency annotation was used, as well as the JOS syntactic dependencies system for Slovenian~\citep{jos-syn}. Additionally, the annotation schema described in~\cite{srl} was used for semantic role label annotation. 

The outline of the model training process given in this paper describes all six tasks supported by CLASSLA-Stanza, however it must be noted that not all tasks are available for every supported language and variety. Dependency parsing has dedicated models for the standard variety of every language except Macedonian. Named entity recognition is also not supported for Macedonian. Processing of the non-standard variety is available only for Slovenian, Croatian and Serbian, while it is not available for Macedonian and Bulgarian. Semantic role labeling currently relies on the JOS annotation system for dependency parsing of Slovenian and is thus only available for annotation of Slovenian datasets, but should become available for Croatian soon, as there is training data available~\citep{hr500k}. Table~\ref{tab:taskslangs} provides an overview of every language variety and the tasks it supports.

\begin{table}[h]
\centering
\begin{tabular}{|c|c|c|c|c|c|c|c|}
\hline
          Language & Variety & Tok & Morph & Lemma & Depparse & NER & SRL \\ \hline \hline
\multirow{2}{*}{Slovenian} & standard  &      \checkmark        &      \checkmark                   &   \checkmark            &   \checkmark                 & \checkmark &    \checkmark                   \\ \cline{2-8}
          & nonstandard &      \checkmark        &      \checkmark                   &   \checkmark            &        X          & \checkmark  &          X           \\ \hline
\multirow{2}{*}{Croatian}   & standard &   \checkmark           &          \checkmark               &       \checkmark        &         \checkmark          & \checkmark &     X                   \\ \cline{2-8}
  & nonstandard &   \checkmark           &          \checkmark               &       \checkmark        &         X          & \checkmark &     X                   \\ \hline
\multirow{2}{*}{Serbian}    & standard &     \checkmark         &           \checkmark              &      \checkmark         &        \checkmark          & \checkmark  &    X                    \\ \cline{2-8}
    & nonstandard &     \checkmark         &           \checkmark              &      \checkmark         &        X           & \checkmark &    X                    \\ \hline
\multirow{2}{*}{Bulgarian}  & standard &     \checkmark         &             \checkmark            &     \checkmark          &         \checkmark          & \checkmark &               X         \\ \cline{2-8}
& nonstandard & X & X & X & X & X & X \\ \hline
\multirow{2}{*}{Macedonian} & standard &     \checkmark         &          \checkmark               &     \checkmark          &        X           & X &              X          \\ \cline{2-8}
& nonstandard & X & X & X & X & X & X \\ \hline
\end{tabular}
\caption{Table illustrating which tasks are supported by CLASSLA-Stanza for every language and variety. The abbreviations for each task are as follows: Tok - tokenization, Morph - morphosyntactic tagging, Lemma - lemmatization, Depparse - dependency parsing, NER - named entity recognition, SRL - semantic role labeling}
\label{tab:taskslangs}
\end{table}

Throughout the paper we use a unified method of providing model evaluation results. For each evaluated model we provide a \textit{Score} value, which summarizes the performance score of that particular model. This value represents different metrics for each task. For lemmatization and semantic role labeling, the \textit{Score} value simply represents the micro F\textsubscript{1} score. The micro F\textsubscript{1} score is calculated by taking the harmonic mean of the precision and recall metrics for all relevant tokens in the dataset. For the morphosyntactic tagging task, the \textit{Score} value represents the micro F\textsubscript{1} score for all three morphosyntactic tagging labels—UPOS, XPOS, and UFeats. Finally, for dependency parsing the \textit{Score} value represents the micro F\textsubscript{1} of the labeled attachment score, or LAS score~\citep{2018-conll}, for every node in the dataset. The LAS score is defined as the percentage of tokens with both a correctly assigned head token as well as a correctly assigned dependency label.

\section{Differences between CLASSLA-Stanza and Stanza}
\label{sec:clasvsstan}

The Stanza neural pipeline is centered around a bidirectional long short-term memory (Bi-LSTM) network architecture~\citep{qi2020stanza}. CLASSLA-Stanza largely preserves the design of Stanza, except in some cases, such as tokenization, where a completely different model architecture is used. CLASSLA-Stanza also expands upon the original design with specific additions that help boost model performance for the South Slavic languages. This section thus lists the main differences between the two pipelines, and in the end provides an overview of the difference in the results produced by the models for one of the supported languages.

On the level of tokenization and sentence segmentation, Stanza uses a joint tokenization and sentence segmentation model based on machine-learning. We generally view such learned tokenizers as suboptimal, since training data for the two tasks is always limited in size and thus too few tokenization and sentence-splitting phenomena can be learned by the model during the training process. Due to this drawback, CLASSLA-Stanza implements rule-based tokenizers, which handle both the task of tokenization as well as sentence segmentation. As stated in the introduction, the two tokenizers used are the Obeliks tokenizer for standard Slovenian~\citep{obeliks} and the ReLDI tokenizer for non-standard Slovenian and all other languages~\citep{reldi-tokenizer}.


CLASSLA-Stanza also adds support for the use of external inflectional lexicons, which is not present in Stanza. For morphologically rich languages, applying this resource to the annotation process usually significantly increases the performance of the model~\citep{ljubesic-dobrovoljc-2019-neural}. The South Slavic languages all have quite rich inflectional paradigms, which is why support for inflectional lexicons is present for almost all supported languages in the pipeline.


Most languages support an external lexicon usage only during lemmatization, except for Slovenian, which supports lexicon use also during morphosyntactic tagging. In that case, the lexicon is put into operation during the tag prediction phase, when the model limits the possible predictions to only those tags that are present in the inflectional lexicon for the specific token. Lexicon usage during lemmatization is similar in both Stanza and CLASSLA-Stanza, the main difference being that Stanza builds a lexicon only from the Universal Dependencies training data, while CLASSLA-Stanza additionally exploits an inflectional lexicon. Both Stanza and CLASSLA-Stanza use the lexicon for an initial lemma lookup, and fall back to predicting the lemma only in case that the form with the corresponding tag is not present in the lexicon. One important difference in the lexicon lookup in CLASSLA-Stanza is that the lookup uses XPOS tags that contain the full morphosynctactic information, while Stanza uses the UPOS tag, which is not enough for an accurate lemma lookup in morphologically rich languages.


When training models, Stanza uses a Universal Dependencies dataset as training data for training all the tasks in the pipeline and thus does not enable the user to train models on additional datasets. For certain layers, however, such as lemmatization and morphosyntactic tagging, the South Slavic languages often have more training data available than for dependency parsing, which is exploited by CLASSLA-Stanza. Thus, for example, instead of using only the 210 thousand tokens of data that are used for training the dependency parser, the latest set of standard Croatian models in CLASSLA-Stanza includes mophosyntactic tagging and lemmatization models which were trained on an additional 290 thousand tokens, which were manually annotated only on these two levels of annotation.


CLASSLA-Stanza also has a special way of handling ``closed-class'' words. Closed-class control is a feature of the tokenizers and ensures that punctuation and symbols are assigned appropriate morphosyntactic tags and lemmas. It also restricts the set of possible tokens that can be assigned morphosyntactic tags and lemmas for punctuation and symbols to only those tokens that are defined as such in the tokenizer. In addition to punctuation and symbols, the Slovenian package also includes closed-class control for pronouns, determiners, adpositions, particles, and coordinating and subordinating conjunctions. These additional closed classes are controlled during the morphosyntactic tagging phase using the inflectional lexicon as a reference, disallowing for any token to be labeled with a closed class label if this token was not defined as such in the lexicon.\footnote{In-depth instructions on how to use the closed-class control functionality are included in the GitHub repository: \url{https://github.com/clarinsi/classla/blob/master/README.closed_classes.md}}



The Stanza pipeline expects pretrained word embeddings as input. While it uses embedding collections based on Wikipedia data, CLASSLA-Stanza does the extra mile by using the CLARIN.SI embeddings~\citep{embed.sl, embed.hr, embed.hr, embed.sr, embed.mk, embed.bg} primarily prepared for CLASSLA-Stanza, but useful for other tasks as well. These embeddings were trained on multiple times larger text collections than Wikipedia that were obtained through web crawling~\citep{macocu_web_crawls}, which ensures drastically more diverse word embeddings and thereby also better unseen word handling.


When working with Slovenian, Croatian, or Serbian, the pipeline can be set to any of three predetermined settings, which are used for processing different varieties of the same language. These settings are called \textit{types} and can be either \textit{standard}, \textit{nonstandard}, or \textit{web}. The processing types determine which model is used on every level of annotation (either standard or nonstandard) and are all associated with their respective language varieties: the \textit{standard} type is used for processing standard language, the \textit{nonstandard} type is used for processing nonstandard Internet language, and the \textit{web} type is used for processing texts obtained from the web. The reasons for introducing a separate processing type for web texts are described in section~\ref{sec:procwebdata}. Below is an overview showing which model is used on every layer for every type:

\begin{table}[h]
	\centering
	\begin{tabular}{|c||c|c|c|c|}
            \hline
		\textbf{Processing type}     & Tokenizer & Morphosyntactic tagger & Lemmatizer & dependency parser \\
		\hline \hline
		\textbf{standard} & standard & standard & standard & standard      \\
            \textbf{nonstandard} & nonstandard & nonstandard & nonstandard & standard      \\
            \textbf{web} & standard & nonstandard & nonstandard & standard      \\
		\hline
	\end{tabular}
        \caption{Overview of processing types in CLASSLA-Stanza}
	\label{tab:proctypes}
\end{table}

The reason why the nonstandard and the web processing type use the standard dependency parsing model is primarily the lack of training data for training a model beyond standard text. The lack of motivation for building a dataset for parsing non-standard text lies in the fact that the parsing model has upstream lemma and morphosyntactic information at its disposal, therefore requires dedicated training data to a much lesser extent than those upstream processes.

To illustrate the performance of CLASSLA-Stanza, Table \ref{tab:slobench_comp} provides a comparison of the results produced by both Stanza and CLASSLA-Stanza when generating predictions on the SloBENCH evaluation dataset. SloBENCH\footnote{\url{https://slobench.cjvt.si/}}~\citep{slobench} is a platform for benchmarking various natural language processing tasks for Slovenian, which includes also a dataset for evaluating the tasks supported by CLASSLA-Stanza. The performance scores are presented in the form of F1 scores, while the relative error reduction between the scores of the pipelines is presented in percentages.

\begin{table}[h]
	\centering
	\begin{tabular}{|c|R{2.8cm}|R{2.8cm}|R{2.8cm}|}
            \hline
		Task & Stanza & CLASSLA-Stanza & Rel. error reduction \\
		\hline \hline
		Sentence segmentation & 0.819 & 0.997 & 98\%     \\
            Tokenization & 0.998 & 0.999 & 50\%     \\
            Lemmatization & 0.974 & 0.992 & 69\%     \\
            Morphosyntactic tagging - XPOS & 0.951 & 0.983 & 65\%     \\
            Dependency parsing LAS & 0.865 & 0.911 & 34\%    \\
		\hline
	\end{tabular}
        \caption{Comparison of performance on the SloBENCH evaluation dataset by both pipelines}
	\label{tab:slobench_comp}
\end{table}

\section{Datasets}
\label{sec:datasets}

The latest models included in the 2.1 release of CLASSLA-Stanza were trained on a variety of datasets in five different languages: Slovenian, Croatian, Serbian, Macedonian, and Bulgarian. Slovenian, Croatian, and Serbian were all associated with two training datasets—one consisting of standard-language texts and one consisting of non-standard texts, while Bulgarian and Macedonian only had a standard-language training dataset available. During training, training datasets for some languages were expanded with additional data due to the size or content of the original datasets. These data augmentation processes are described in more detail in section \ref{sec:langspecproc}.

Slovenian standard language models were trained using the SUK training corpus~\citep{SUK}. It contains approximately 1 million tokens of text manually annotated on the levels of tokenization, sentence segmentation, morphosyntactic tagging, and lemmatization. Some subsets also contain syntactic dependency, named entity, multi-word expression, coreference, and semantic role labeling annotations. The corpus is a continuation of the ssj500k Slovene training corpus~\citep{ssj500k}. Non-standard models were trained using the Janes-Tag training corpus~\citep{janes-tag}, which consists of tweets, blogs, forums, and news comments, and is approximately 218 thousand tokens in size. It contains manually curated annotations on the levels of tokenization, sentence segmentation, word normalization, morphosyntactic tagging, lemmatization, and named entity annotation.

Croatian standard language models were trained on the hr500k training corpus~\citep{hr500k}, which consists of about 500 thousand tokens and is manually annotated on the levels of tokenization, sentence segmentation, morphosyntactic tagging, lemmatization, and named entities. Portions of the corpus also contain manual syntactic dependency, multi-word expression, and semantic role labeling annotations. Croatian non-standard models were trained using the ReLDI-NormTagNER-hr training corpus~\citep{reldi-normtagner-hr}. It contains about 90 thousand tokens of non-standard Croatian text from tweets and is manually annotated on the levels of tokenization, sentence segmentation, word normalization, morphosyntactic tagging, lemmatization, and named entity recognition.

Serbian standard models were trained on the Serbian portion of the SETimes corpus~\citep{setimes-sr}, which contains about 97 thousand tokens of news articles manually annotated on the levels of tokenization, sentence segmentation, morphosyntactic tagging, lemmatization, and dependency parsing.
Serbian non-standard models were trained using the ReLDI-NormTagNER-sr training corpus~\citep{reldi-normtagner-sr}. It consists of about 90 thousand tokens of Serbian tweets manually annotated on the levels of tokenization, sentence segmentation, word normalization, morphosyntactic tagging, lemmatization, and named entity recognition.

Macedonian standard models were trained on a corpus made up of the Macedonian version of the MULTEXT-East ``1984'' corpus~\citep{1984} and the Macedonian portion of the SETimes corpus. The MULTEXT-East ``1984'' corpus consists of the novel \textit{1984} by George Orwell in approximately 113 tousand tokens, while the SETimes corpus is currently made up of about 13 thousand tokens of news articles. At the time of writing this paper, neither of the datasets have unfortunately not yet been made publicly available, as they are still in the process of being published. Both corpora are manually annotated on the levels of tokenization, sentence segmentation, morphosyntactic tagging, and lemmatization. The combining of the corpus was performed in the following way: the 1984 corpus was first split into three parts to obtain the training, validation and testing data splits, after which only the training data split was enriched with three repetitions of the SETimes corpus to ensure a sensible combination of literary and newspaper data in the training subset.

Bulgarian standard models were trained on the BulTreeBank training corpus~\citep{bultreebank}, which consists of approximately 253 tousand tokens manually annotated on the levels of tokenization, sentence segmentation, morphosyntactic tagging, and lemmatization. About 60~\% of the dataset also contains manual dependency parsing annotations.

Table~\ref{tab:traindata} provides an overview of dataset sizes for every language, variety, and annotation layer. 

\begin{table}[h]
\centering
\begin{tabular}{|c|c||c|c|c|c|}
\hline
          Language & Variety & Morph & Lemma & Depparse & SRL \\ \hline \hline
\rule{0pt}{10pt}\multirow{2}{*}{Slovenian} & standard        &      1,025,639                   &   1,025,639            &   267,097                 &     209,791                   \\ \cline{2-6}
\rule{0pt}{10pt}      & nonstandard      &      222,132                   &   222,132            &        n/a            &          n/a           \\ \hline
\rule{0pt}{10pt}\multirow{2}{*}{Croatian}   & standard       &          499,635               &       499,635        &         199,409           &     n/a                   \\ \cline{2-6}
\rule{0pt}{10pt}  & nonstandard &          89,855               &       89,855        &         n/a           &     n/a                   \\ \hline
\rule{0pt}{10pt}\multirow{2}{*}{Serbian}    & standard         &           97,673              &      97,673         &        97,673            &    n/a                    \\ \cline{2-6}
\rule{0pt}{10pt}    & nonstandard &           92,271              &      92,271         &        n/a            &    n/a                    \\ \hline
\rule{0pt}{10pt}Bulgarian  & standard        &             253,018           &     253,018          &         156,149           &               n/a         \\\hline
\rule{0pt}{10pt}Macedonian & standard        &          153,091               &     153,091          &        n/a            &              n/a          \\\hline
\end{tabular}
\caption{Overview table of the number of tokens annotated on every annotation layer for all training datasets used. The abbreviations for each task are as follows: Morph - morphosyntactic tagging, Lemma - lemmatization, Depparse - dependency parsing, SRL - semantic role labeling}
\label{tab:traindata}
\end{table}

\section{Model Training Process}
\label{sec:modtrapro}

In this section, the model training process is described in detail. Only a descriptive account of the process is provided here. For a list of the specific commands and oversampling scripts used, refer to the GitHub repository of the training procedure\footnote{\url{https://github.com/clarinsi/classla-training}}.

In section \ref{sec:generalproc}, a general overview of the process which is common to all supported languages is first given. In section \ref{sec:langspecproc}, the specific steps that are unique to each language are then described. This structuring is due to some features and levels of annotation (semantic role labeling, oversampling of the training data, etc.) being unique to only certain languages, while all languages at the same time share a few common steps.

\subsection{General Procedure}
\label{sec:generalproc}

As stated in the introduction, all tokenizers used by CLASSLA-Stanza are rule-based and thus do not need to be trained. Model training is thus performed on pretokenized data, typically beginning on the level of morphosyntactic tagging and continuing on through the subsequent annotation layers.

To ensure realistic evaluation results, automatically generated upstream annotations, rather than manually assigned annotations, were used as validation and test dataset inputs on each layer. For this, empty validation and test datasets first had to be generated by stripping all annotations from the test and validation datasets on all levels except for tokenization. These empty files were filled with model-generated annotations on each level, so that validation and model evaluation on subsequent layers could be performed on automatically generated upstream labels. Training datasets were not annotated with automatically generated upstream labels, since it is unclear whether this would lead to any performance gains and would require a more complicated type of cross-validation method such as jackknifing (splitting data into N bins, training a model on N-1 bins and annotating the N-th bin, repeating the process N times).


For each language, standard models were first trained. For morphosyntactic tagging training, the training and validation datasets from the prepared three-way data split along with the pretrained word embeddings were used as inputs to the tagger module. After training, the tagger was used in predict mode to generate predictions on the empty test dataset and evaluate the performance of the tagger. After predictions were made for the test set, predictions were generated in the empty validation dataset as well, so as to produce a validation file with automatically generated upstream labels, that can be used later during training of subsequent annotation layer models.

Once morphosyntactic predictions and evaluation results were obtained, the lemmatizer was trained. The validation and training datasets were used as inputs. In addition, for most languages, the inflectional lexicon is also provided to the lemmatizer as one of the inputs. During training, the lexicon is stored in the lemmatization model file to act as an additional controlling element during lemmatization. After training, the lemmatizer was run in predict mode to obtain evaluation results and add lemma predictions to the validation and test datasets.

The dependency parser module was trained after lemmatizer training was finalized. CLASSLA-Stanza currently supports two types of annotation systems for syntactic dependency annotation: the UD dependency parsing annotation system, which is available for all supported languages except Macedonian, and the JOS parsing system, which is only available for Slovenian. The parser was run in training mode using the training and validation datasets\footnote{For most languages, only a portion of the original datasets contained dependency parsing annotations. In these cases, a separate set of training, validation, and test datasets consisting of only this portion of the original data had to be extracted.} as inputs along with the pretrained word embeddings. After training, the parser was run in predict mode to obtain evaluation results and add automatically generated dependency parsing annotations to the validation and test datasets.

For this latest release of CLASSLA-Stanza, no new models for named-entity recognition were trained. However, the process for training models for named-entity recognition is quite similar to the other tasks. The tagger trainer for this task accepts pretrained word embeddings and training and validation datasets as inputs. After training, the named-entity recognition tagger can be run in predict mode to obtain evaluation results and add model predictions to the relevant datasets. 

The non-standard models were trained using the same process as the standard models, with a few exceptions: firstly, no syntactic dependency annotations were present in the non-standard datasets. As a result, no non-standard dependency parsing models were trained. On each level of annotation, both the non-standard as well as the standard test datasets were used in order to gain a good understanding of how well the model performs on different varieties. Additionally, before training the non-standard models, approximately 20~\% of diacritics were removed from the training datasets in order to ensure that the models will learn to effectively handle dediacritized forms, which occur prominently in online communication. It is important to note that the non-standard data was regularly oversampled. Whenever this was done, care was taken not to remove the diacritics only from copies of an oversampled dataset. This method ensured that no global information loss would result from the diacritic removal process.


\subsection{Language-specific Procedure}
\label{sec:langspecproc}

Languages vary with respect to the amount of deviation from the general procedure. We describe the training process for each language in its own subsection below.

\subsubsection{Slovenian}


In CLASSLA-Stanza, Slovenian differs from other languages in the fact that the inflectional lexicon is applied as input during the training of the morphosyntactic tagger and not of the lemmatizer, as is the case with all other languages. The lexicon is stored within the morphosyntactic tagger model file and can then be used on both the level of morphosyntactic tagging and lemmatization. The Sloleks 3.0 inflectional lexicon~\citep{sloleks} was used during the training of the latest Slovenian models. Important to add is that the lexicon was added to the tagger only for the standard variety models as for the non-standard Internet variety the lexicon control during tagging proved to be detrimental. Non-standard models were set-up as for all other languages with the lemmatizer storing and using the external inflectional lexicon only.

For Slovenian, CLASSLA-Stanza also supports semantic role labeling (SRL). This module relies on the JOS dependency parsing system, thus JOS syntactic annotations must be present in the training, validation and test datasets which are used as inputs to the semantic role tagger. Pretrained word embeddings must also be provided to the SRL tagger during training.

\begin{figure}[h]
\centering
\includegraphics[scale=0.7]{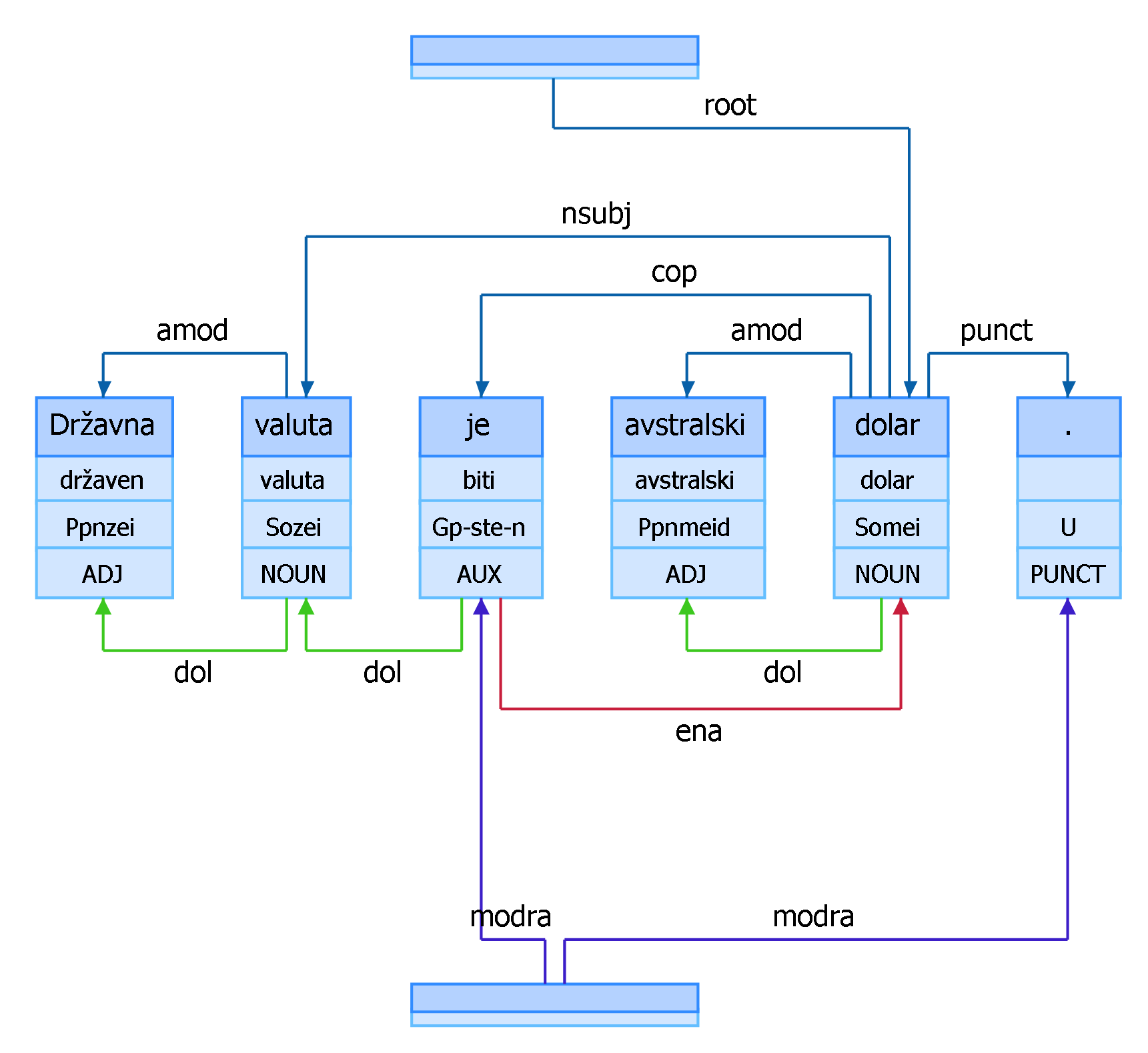}
\caption{Example of a sentence parsed with both the UD (above) and JOS (below) systems of syntactic annotation. Note that the JOS system supports multiple relations having the root element as their origin, whereas the UD annotation system only permits one relation in each sentence to originate in the root element.} \label{fig:jos-ud}
\end{figure}

Another way in which Slovenian differs from other languages is that, in addition to the UD dependency parsing annotation system, the JOS syntactic annotation system is also supported. An example of a Slovenian sentence that is parsed using both systems is shown in Figure \ref{fig:jos-ud}. The training procedure for this system is identical to UD dependency parsing, with the exception that the option for allowing multiple roots in syntactic trees must be passed to the parser during training.

After training the models for all the annotation layers listed above, the resulting models for standard Slovenian yielded the results in Table \ref{tab:slostan}. The relationship in performance of different layers is similar to the one reported for the Stanza pipeline in \cite{qi2020stanza}, where the highest performance is generally displayed by the lemmatizer, followed by the morphosyntactic tagger, and then by the dependency parser. The performance scores of the models correlate with the amount of training data available for each layer, except in the case of both syntactic parsers, which were trained on the same training dataset. An important cause for the lower performance score of the UD syntactic parser may be the higher granularity of the UD dependency annotation scheme compared to the 10 dependency relation labels used by the JOS dependency annotation scheme. The semantic role labeling model displays the lowest performance scores of all with a significant performance drop compared to the other tasks, suggesting that the current amount of available training data is not sufficient for learning the abstract patterns involved in the task.

\begin{table}[h]
	\centering
	\begin{tabular}{|c|c|}
            \hline
		Model     & Score \\
		\hline \hline
		Morphosyntactic tagger & 97.08      \\
            Lemmatizer & 98.97      \\
            UD syntactic parser & 90.57      \\
            JOS syntactic parser & 93.89      \\
            Semantic role labeling & 76.24      \\
		\hline
	\end{tabular}
        \caption{Overview of Slovenian standard model performance}
	\label{tab:slostan}
\end{table}

Before training the non-standard models, a series of pre-processing operations had to be made on the non-standard dataset. The Janes-Tag 3.0 non-standard training corpus~\citep{janes-tag} contains also forms which were originally written as one word in non-standard communication, but correspond to several separated words in standard Slovenian. During the tokenization process, the original merged form was preserved in the corpus and the standardized separated forms were added to the dataset, clearly stating their relationship through token indices. An example of this phenomenon as it is encoded in a CoNLL-U formatted file is shown below. Only the combined form \textit{tastare} appeared in the original sentence, the two separated subtokens \textit{ta} and \textit{stare} were added during manual annotation of the Janes-Tag corpus:

\begin{alltt}
    # sent_id = tid.1265356974629228544.s2
    # text = mislm sej mam zdej shrambo polno pa tud notr so bli nakupi za tastarezihr par 
    100e sam vseen me je mal kap.
    [...]
    [...]
    12 nakupi _ _ _ _ _ _ _ _
    13 za _ _ _ _ _ _ _ _
    \textbf{14-15 tastare _ _ _ _ _ _ _ _
    14 ta _ _ _ _ _ _ _ _
    15 stare _ _ _ _ _ _ _ _}
    16 zihr _ _ _ _ _ _ _ _
    17 par _ _ _ _ _ _ _ _
    18 100 _ _ _ _ _ _ _ _
    19 e _ _ _ _ _ _ _ _
\end{alltt}

Similarly the corpus also contains forms which are separated in non-standard communication, but should be written as one word in standard Slovenian. In the example below the word \textit{parlament} was written as two separate words in the original text. The two separated words along with the space character are treated as one single token in the CoNLL-U formatted file:

\begin{alltt}
    # sent_id = tid.892838763793182720.s1
    # text = @leaathenatabako hahahahaha ja maš mrbit parla ment mau bliži
    [...]
    [...]
    5 mrbit _ _ _ _ _ _ _ _
    \textbf{6 parla ment _ _ _ _ _ _ _ _}
    7 mau _ _ _ _ _ _ _ _
    8 bliži _ _ _ _ _ _ _ _
\end{alltt}

These two special cases of tokenization are referred to as 1:n and n:1 cases. In order for the data to be properly processed by CLASSLA-Stanza, the spaces in the n:1 cases had to be removed, essentially merging the constituent parts of the multi-token unit. The example above thus had to be reduced from ``parla ment'' to a single word \textit{parlament}. The 1:n cases, on the other hand, did not require any alteration, since CLASSLA-Stanza currently ignores the composite forms and trains on the separated forms. This is not an ideal solution, as the model should be trained on the non-standard forms as they appear in the original text and not on the normalized forms as it currently does. This limitation stems from the fact that the currently employed tokenizers were not set up to split or merge words.

After handling the non-standard tokenization issues, the Slovenian non-standard and standard training datasets were merged into a single combined training dataset. This was done because social media or web text is expected to consist of a substantial amount of standard forms and the model is expected to perform better on these if it is trained on a combined dataset. However, since the standard training dataset for Slovenian is quite notably larger than the non-standard dataset, some amount of oversampling had to be applied. In order to achieve a 1:1 ratio of standard to non-standard data in the resulting combined dataset, 4.7 repetitions of the non-standard data were combined with a single repetition of the standard data. Additionally, in 1.5 of the non-standard data repetitions, all diacritics were removed to ensure that approximately 20 \% of all forms in the training dataset were dediacritized.

Once the data preparation step was completed, the non-standard models for morphosyntactic tagging and lemmatization were trained and afterwards evaluated on both the standard and non-standard test datasets. The evaluation results are presented in Table \ref{tab:slononst}. The non-standard models perform somewhat better on standard language than on non-standard language, however this is to be expected to a certain degree due to the less regular nature of non-standard language. The same pattern of slightly better performance of the lemmatizer compared to the morphosyntactic tagger, which can be observed in the standard models, is preserved in the non-standard models as well. 

\begin{table}[h]
	\centering
	\begin{tabular}{|c|c|c|}
            \hline
		Model  & Test dataset   & Score \\
		\hline \hline
		Morphosyntactic tagger & non-standard & 91.51     \\
            Morphosyntactic tagger & standard & 95.80     \\
            Lemmatizer & non-standard & 91.63     \\
            Lemmatizer & standard & 98.77     \\
		\hline
	\end{tabular}
        \caption{Overview of Slovenian non-standard model performance}
	\label{tab:slononst}
\end{table}

\subsubsection{Croatian}

The models for standard Croatian were trained with no special digressions from the general procedure. Standard Croatian models were produced for morphosyntactic tagging, lemmatization, and dependency parsing. The hrLex 1.3 inflectional lexicon~\citep{hrlex} was used for building the lemmatizer. The results are displayed in Table \ref{tab:crostan}.

\begin{table}[h]
	\centering
	\begin{tabular}{|c|c|}
            \hline
		Model     & Score \\
		\hline \hline
		Morphosyntactic tagger & 94.33      \\
            Lemmatizer & 98.02      \\
            UD syntactic parser & 87.46      \\
		\hline
	\end{tabular}
        \caption{Overview of Croatian standard model performance}
	\label{tab:crostan}
\end{table}

Similarly to Slovenian, the Croatian non-standard models were also trained on a combination of standard and non-standard training data. In the case of Croatian, 5.5 repetitions of the non-standard training dataset and a single instance of the standard training dataset were needed to construct a combined dataset with a 1:1 ratio of standard to non-standard data. Additionally, diacritics were removed from two whole repetitions of the non-standard training dataset to arrive at roughly 20\% of data with removed diacritics. Evaluation was performed on the standard and non-standard test datasets. The results are shown in Table \ref{tab:crononst}

\begin{table}[h]
	\centering
	\begin{tabular}{|c|c|c|}
            \hline
		Model  & Test dataset   & Score \\
		\hline \hline
		Morphosyntactic tagger & non-standard & 91.09     \\
            Morphosyntactic tagger & standard & 94.03     \\
            Lemmatizer & non-standard & 93.61     \\
            Lemmatizer & standard & 97.68     \\
		\hline
	\end{tabular}
        \caption{Overview of Croatian non-standard model performance}
	\label{tab:crononst}
\end{table}

We observe similar phenomena as with Slovenian non-standard models. Non-standard models perform slightly worse than standard models on standard test data due to higher variation in the training data themselves. Performance on non-standard test data is especially lower because of the higher variation in the test data. We consider the similarity of results across languages to be a positive signal for the robustness of the specific solutions inside this processing pipeline.

\subsubsection{Serbian}

Due to the relatively small size of the Serbian training datasets in comparison to the Croatian datasets, and the fact that the two languages are very similar, the training data for Serbian had to be prepared in a special way for each task. On the level of morphosyntactic tagging the Serbian data were combined with Croatian data in a specific ratio. This method of data augmentation is possible due to the very high similarity of the two languages and previous experimental results showing very good performance on Serbian test data if Croatian training data were used~\citep{ljubesic-etal-2016-new}. However, combining Serbian and Croatian data could be performed only for the task of morphosyntactic tagging and not lemmatization as well, since words are lemmatized quite differently in the two languages. While Serbian prefers the ekavian variant, Croatian mostly exclusively uses the ijekavian one~\citep{ljubevsic2018borders}, often resulting in different lemmas. The SrLex 1.3 inflectional lexicon~\citep{srlex} was used during the training of the lemmatizer.

Models for the processing of standard Serbian were trained first. In the case of morphosyntactic tagging, 5.4 repetitions of the Serbian standard training dataset were combined with a single instance of the Croatian standard training dataset, thus reaching a 1:1 ratio of Serbian to Croatian data. For lemmatization, the Serbian standard and non-standard datasets were merged. No data augmentation was applied for dependency parsing. The models were trained in accordance with the general procedure, after which the evaluation produced the results shown in Table~\ref{tab:serstan}.

\begin{table}[h]
	\centering
	\begin{tabular}{|c|c|}
            \hline
		Model     & Score \\
		\hline \hline
		Morphosyntactic tagger & 95.79      \\
            Lemmatizer & 98.02      \\
            UD syntactic parser & 89.83      \\
		\hline
	\end{tabular}
        \caption{Overview of Serbian standard model performance}
	\label{tab:serstan}
\end{table}

In order to construct the training dataset for the Serbian non-standard morphosyntactic tagger, the standard Serbian training dataset was first expanded with a similar amount of data from the Croatian standard training dataset. While sampling from the Croatian standard training dataset preference was given to sentences that do not originate from the SETimes part of the Croatian standard dataset as the Serbian standard dataset consists mostly of sentences from the Serbian translation of the same SETimes dataset. In accordance with this, about 22 \% of the non-SETimes portion of the standard Croatian dataset was required. This combined standard dataset was then merged with two repetitions of the non-standard Serbian training dataset, with all diacritics removed from one of the repetitions, so as to maintain the 1:1 ratio of standard to non-standard data.

The training data for Serbian non-standard lemmatization was handled in a similar way to the standard lemmatizer training data. Equal amounts of Serbian standard and non-standard training data were combined into a single dataset. However, since the Serbian standard and non-standard training datasets are about the same in size to begin with, no repetition of data was necessary. From that point, if any dediacritization of data would have been performed, it would have resulted in loss of original data. To avoid this, no dediacritization was performed for the non-standard lemmatization training dataset. This decision was backed up by the fact that some dediacritized forms are already present in non-standard online texts and the model could learn at least some patterns of dediacritized form usage without any additional modifications of the data.

Evaluation of the non-standard models was performed both on the standard and non-standard testing datasets. The results are displayed in Table~\ref{tab:sernonst}.

\begin{table}[h]
	\centering
	\begin{tabular}{|c|c|c|}
            \hline
		Model  & Test dataset   & Score \\
		\hline \hline
		Morphosyntactic tagger & non-standard & 92.56     \\
            Morphosyntactic tagger & standard & 95.32     \\
            Lemmatizer & non-standard & 94.92     \\
            Lemmatizer & standard & 98.02     \\
		\hline
	\end{tabular}
        \caption{Overview of Serbian non-standard model performance}
	\label{tab:sernonst}
\end{table}

In both the evaluations of the standard and the non-standard model the same relationships can be observed as was the case with Slovenian and Croatian, which is a very reassuring signal in the stability and robustness of the presented solutions.

\subsubsection{Macedonian}

The Macedonian models were trained largely in accordance with the general procedure, with only a few alterations: only models for morphosyntactic tagging and lemmatization were trained, as no dependency parsing data was available at time of training. Additionally, no inflectional lexicon was used during lemmatization, and models were trained only for the standard variety.

Model evaluation was performed on the testing dataset. The results are displayed in Table~\ref{tab:mkstan}.

\begin{table}[h]
	\centering
	\begin{tabular}{|c|c|c|}
            \hline
		Model     & Score \\
		\hline \hline
		Morphosyntactic tagger &  96.99     \\
            Lemmatizer &  98.81     \\
		\hline
	\end{tabular}
        \caption{Overview of Macedonian standard model performance}
	\label{tab:mkstan}
\end{table}

The rather high evaluation metrics are due to the fact that Macedonian testing data come from the ``1984'' dataset, which has limited diversity, and is heavily present in the training data as well.

\subsubsection{Bulgarian}

As with Macedonian, the Bulgarian training data also only consists of standard-language texts, thus no non-standard models were trained for this language. Models were trained for the tasks of morphosyntactic tagging, lemmatization, and dependency parsing. A smaller portion of the Bulgarian dataset was used to train the dependency parser, as only this part contains dependency parsing annotations. The \textit{Dictionary of Writing, Pronunciation and Punctuation of Bulgarian Language}~\citep{buldict} was used as the inflectional lexicon during lemmatization training.

Model evaluation was performed on the testing dataset. The results are displayed in Table~\ref{tab:bgstan}.

\begin{table}[h]
	\centering
	\begin{tabular}{|c|c|}
            \hline
		Model     & Score \\
		\hline \hline
		Morphosyntactic tagger & 94.46      \\
            Lemmatizer & 98.93      \\
            UD syntactic parser & 91.18      \\
		\hline
	\end{tabular}
        \caption{Overview of Bulgarian standard model performance}
	\label{tab:bgstan}
\end{table}

The evaluation metrics follow very well the evaluation results of the previously evaluated languages.

\section{Model Performance Analysis}
\label{sec:erran}

The performance scores obtained during the model evaluations give a concise overview of how well CLASSLA-Stanza handles each task. The distribution of the results across the various annotation tasks are fairly consistent between the languages and their varieties. In addition, as noted in section \ref{sec:clasvsstan}, CLASSLA-Stanza significantly outperforms Stanza on the Slovenian example, with error reduction between 34\% and 98\%. 

However, in order to fully assess the performance of the newly-trained models, a series of additional performance analyses was performed. In section \ref{sec:specificlabels} a detailed rundown of the performance of the models for specific labels is given. Section \ref{sec:procwebdata} then continues with an investigation of the performance of the models on web data.

\subsection{Model Performance on Specific Labels}
\label{sec:specificlabels}

The evaluation carried out after training each model proved useful to create an overall impression of the performance. However, it is also useful to identify which specific categories a model struggles with and which ones it handles with particular ease. To obtain a sense of such performance patterns, model predictions for specific UPOS and UD syntactic relations were inspected. An accuracy score was calculated for all 17 UPOS labels and the 12 most frequent UD syntactic relations in the Croatian hr500k training corpus. The accuracy score was obtained by taking the number of correct predictions for a single label in the test dataset and dividing it by the total number of occurrences of that label in the test dataset. The resulting accuracies for all the UPOS tags are contained in Table \ref{tab:uposperrel}, while Table \ref{tab:deprelperrel} contains accuracies for each UD dependency relation.

\begin{table}[h]
    \centering
    \begin{tabular}{|c|c|c|c|c|c|c|c|c||c|}
			\hline
			\multirow{2}{*}{UPOS tag} & \multicolumn{9}{|c|}{Accuracy} \\
                \cline{2-10}
                 & sl-st & sl-nonst & hr-st & hr-nonst & sr-st & sr-nonst & mk-st & bg-st & Average \\
			\hline \hline
    		ADJ & 99.31 & 90.71 & 97.93 & 92.27 & 99.27 & 94.58 & 97.74 & 98.28 & 96.26 \\
                \hline
                ADP & 99.90 & 98.54 & 99.96 & 99.82 & 100.00 & 99.84 & 99.75 & 99.92 & 99.72 \\
                \hline
                ADV & 95.98 & 91.89 & 95.35 & 91.59 & 95.42 & 87.93 & 95.14 & 97.60 & 93.86 \\
                \hline
                AUX & 98.62 & 96.31 & 99.60 & 99.59 & 100.00 & 98.81 & 99.50 & 92.75 & 98.15 \\
                \hline
                CCONJ & 98.01 & 97.03 & 96.53 & 97.21 & 98.95 & 97.21 & 97.94 & 97.87 & 97.59 \\
                \hline
                DET & 99.29 & 93.29 & 95.68 & 94.08 & 98.88 & 96.74 & 100.00 & 87.79 & 95.72 \\
                \hline
                INTJ & 80.00 & 75.82 & 71.43 & 90.22 & n/a & 87.65 & 71.43 & 47.58 & 74.88 \\
                \hline
                NOUN & 98.88 & 93.75 & 98.33 & 93.98 & 99.23 & 97.66 & 99.55 & 98.53 & 97.49 \\
                \hline
                NUM & 99.74 & 98.41 & 98.87 & 100.00 & 98.71 & 100.00 & 100.00 & 98.17 & 99.24 \\
                \hline
                PART & 99.46 & 95.12 & 85.16 & 90.64 & 94.12 & 89.39 & 90.16 & 79.94 & 90.50 \\
                \hline
                PRON & 99.47 & 97.25 & 98.68 & 98.19 & 97.64 & 98.47 & 98.84 & 99.15 & 98.46 \\
                \hline
                PROPN & 98.71 & 78.23 & 93.65 & 77.81 & 97.31 & 83.68 & 97.97 & 98.14 & 90.69 \\
                \hline
                PUNCT & 100.00 & 99.79 & 100.00 & 99.73 & 100.00 & 99.82 & 100.00 & 100.00 & 99.92 \\ 
                \hline
                SCONJ & 99.78 & 97.99 & 95.72 & 94.79 & 99.52 & 98.25 & 94.70 & 99.61 & 97.55 \\
                \hline
                SYM & 100.00 & 99.85 & 90.91 & 99.10 & 100.00 & 99.38 & n/a & n/a & 98.21 \\
                \hline
                VERB & 97.05 & 94.12 & 99.30 & 97.84 & 99.18 & 98.76 & 99.74 & 96.79 & 97.85 \\
                \hline
                X & 59.13 & 75.67 & 77.15 & 80.10 & 43.33 & 62.86 & n/a & 0.00 & 56.89 \\
			\hline
		\end{tabular}
    \caption{Table of per-relation accuracies for all UPOS tags. The language abbreviations are followed by either ``st'' for \textit{standard} or ``nonst'' for \textit{non-standard}}
    \label{tab:uposperrel}
\end{table}

\begin{table}[h]
    \centering
    \begin{tabular}{|c|c|c|c|c||c|}
			\hline
			\multirow{2}{*}{UD relation} & \multicolumn{5}{|c|}{Accuracy} \\
                \cline{2-6}
                 & sl & hr & sr & bg & Average \\
			\hline \hline
			punct & 100.00 & 100.00 & 100.00 & 99.91 & 99.98 \\
                \hline
                amod & 98.61 & 95.97 & 97.38 & 98.66 & 97.66 \\
                \hline
                case & 99.63 & 99.32 & 99.21 & 99.86 & 99.51 \\
                \hline
                nmod & 92.74 & 91.22 & 90.99 & 91.49 & 91.61 \\
                \hline
                nsubj & 90.49 & 93.39 & 94.30 & 91.10 & 92.32 \\
                \hline
                obl & 91.99 & 85.31 & 87.24 & 77.17 & 85.43 \\
                \hline
                conj & 92.51 & 90.92 & 93.06 & 93.95 & 92.61 \\
                \hline
                root & 93.14 & 94.98 & 95.77 & 95.97 & 94.97 \\
                \hline
                obj & 93.33 & 82.84 & 91.39 & 90.18 & 89.44 \\
                \hline
                aux & 99.48 & 97.88 & 97.57 & 90.46 & 96.35 \\
                \hline
                cc & 97.83 & 97.63 & 97.96 & 99.14 & 98.14 \\
                \hline
                advmod & 96.74 & 93.58 & 91.82 & 97.91 & 95.01 \\
			\hline
		\end{tabular}
    \caption{Table of per-relation accuracies for all UD relations}
    \label{tab:deprelperrel}
\end{table}

The highest accuracies among UPOS tags are generally found with tags that represent function word classes, such as \textbf{AUX} (auxiliaries), \textbf{ADP} (adpositions), and \textbf{PRON} (pronouns), and closed-class tags, such as \textbf{PUNCT} (punctuation) and \textbf{SYM} (symbols), which are handled by the pipeline, inter alia, through rules in the tokenizer, as described in section \ref{sec:clasvsstan}. Conversely, the lowest accuracies are found with the infrequent \textbf{INTJ} tag (interjections)—of which there were only 5 instances in total in the Slovenian standard test dataset and no instances at all in the Serbian standard test dataset—and the loosely delineated \textbf{X} tag, which is used for abbreviations, URLs, foreign language tokens, and everything else that does not fit into any of the other categories. 

A similar trend is found among the UD syntactic relations. Relations such as \textbf{case} (which usually connects nominal heads with adpositions), \textbf{cc} (connects conjunct heads with coordinating conjunctions), and \textbf{aux} (connects verbal heads with auxiliary verbs) are used for fixed grammatical patterns that permit little variation. These display consistently high accuracies across all languages. Somewhat lower accuracies are displayed by the \textbf{obl} relation, mostly used for oblique nominal arguments, which play a less central role in the sentence structure than the core verbal arguments. It has been found that previous versions of dependency parsing models for CLASSLA-Stanza often incorrectly assigned the \textbf{obj} relation (used for direct objects) to instances which should receive the \textbf{obl} relation and vice versa~\citep{jtdh-ud}. Upon inspection of the outputs produced by the newly-trained Slovenian and Croatian parsers it was found that this error persists also in the current version, which is also a likely reason for the performance drops of the \textbf{obl} and \textbf{obj} relations in other languages as well.  

\subsection{Model Performance on Web Data}
\label{sec:procwebdata}

The model evaluations described in previous sections provide a good summary of how well the CLASSLA-Stanza pipeline performs on both purely standard and purely non-standard data. However, modern corpus construction techniques—especially for low-resource languages—often rely on crawling data from online conversations, articles, blogs, etc.~\citep{corpuscoll}, which typically consists of a mixture of different language styles and varieties. To illustrate how well the new CLASSLA-Stanza models handle language originating on the internet, this section provides a brief overview of their performance on a corpus of web data.

The CLASSLA-Stanza tool was used with the newly-trained models to add linguistic annotations to the CLASSLA-web corpora, which consist of texts crawled from the internet domains of the corresponding languages~\citep{macocu-sl, macocu-hr}. In preparation for the annotation process a short test was conducted with the goal of determining which of the two sets of models—the standard or the non-standard—is best suited to be used for annotating the CLASSLA-web corpora. Shorter portions of the corpora were annotated on the levels of tokenization, sentence segmentation, morphosyntactic tagging and lemmatization, once using the standard and once using the non-standard models. The two outputs were then compared and a qualitative analysis of the differences was conducted.

Quite a few of the analyzed differences in the model outputs were connected to the processes of sentence segmentation and tokenization. In the CLASSLA-Stanza annotation pipeline, both of these processes are controlled by the tokenizer. As stated in section \ref{sec:clasvsstan}, the pipeline uses two different tokenizers depending on the language and the annotation type used~\footnote{The ReLDI tokenizer can be used in two different settings: standard and non-standard. The Obeliks tokenizer, on the other hand, only supports tokenization of standard text}. The analysis showed that sentence segmentation was performed much more accurately by Obeliks and the standard mode of the ReLDI tokenizer. The non-standard mode of the ReLDI tokenizer appears to have a tendency towards producing shorter segments, since it is optimized for processing social media texts such as tweets. Thus, the non-standard tokenizer very consistently produces a new sentence after periods, question marks, exclamation marks, and other punctuation, even when these characters do not signify the end of a segment. The following Croatian example in a simplified CoNLL-U format shows one such case of incorrect sentence segmentation, due to the use of reported speech. The original string \textit{„ Svaku našu riječ treba da čuvamo kao najveće blago.“} was split into two segments - the first ending on the period character, while the quotation mark was moved to a separate sentence:

\begin{alltt}
    # newpar id = 76
    # sent_id = 76.1
    # text = „ Svaku našu riječ treba da čuvamo kao najveće blago.
    1 „
    2 Svaku
    3 našu
    4 riječ
    5 treba
    6 da
    7 čuvamo
    8 kao
    9 najveće
    10 blago
    11 .
    
    # sent_id = 76.2
    # text = “
    1 “
\end{alltt}



Besides sentence segmentation issues, the standard models also performed better than the non-standard models when assigning certain types of grammatical features, such as with disambiguating between the UD part-of-speech labels AUX and VERB for the verb \textit{biti} (Eng. ``to be''). However, the difference between the two model outputs for these grammatical features was not as noticeable as on the levels of tokenization and sentence segmentation.

The non-standard models, on the other hand, handled non-standard word forms quite a bit better than the standard models. Particularly problematic for the standard Slovenian models were forms with missing diacritics, such as ``sel'' instead of \textit{šel}, ``cist'' instead of \textit{čisto}, ``hoce'' instead of \textit{hoče}, and ``clovek'' instead of \textit{človek}. These were often assigned incorrect lemmas and morphosyntactic tags. An example of the standard lemmatizer output for the word form ``hoce'' (which corresponds to \textit{hoče} in standard Slovene (Eng. ``he/she/it wants'')) is displayed below. The model invents a nonexistent lemma ``hocati'', while the correct form should be the standard Slovenian \textit{hoteti}:

\begin{alltt}
    # sent_id = 53.1
    # text = lev je lev pa naj govori kar kdo hoce
    1 lev lev
    2 je biti
    3 lev lev
    4 pa pa
    5 naj naj
    6 govori govoriti
    7 kar kar
    8 kdo kdo
    \textbf{9 hoce hocati}
\end{alltt}


Non-standard forms which do not differ much from their standard counterparts, such as ``zdej'' as opposed to ``zdaj'' and ``morš'' as opposed to ``moraš'', were generally handled well by both sets of models and did not cause many discrepancies in the outputs.

The analysis of such differences in the model outputs showed that the best results for the web corpus were achieved on the one hand by the standard tokenizer, and on the other by the non-standard models for all subsequent levels of annotation. In light of this, a new \textit{web} type was implemented for the CLASSLA-Stanza pipeline. This new type combines the standard tokenizer and non-standard models for the other layers in a single package and is intended specifically for the annotation of texts originating on the internet.



\section{Conclusion}


In this paper we provided an overview of the CLASSLA-Stanza pipeline for linguistic processing of the South Slavic languages and described the training process for the models included in the latest release of the pipeline. We described the main design differences to the Stanza neural pipeline from which CLASSLA-Stanza arose as a forked project. We provided a summary of the model training process, first in a general outline and then in a more detailed procedure description for each supported language. We also presented performance scores for each model trained for the latest release of CLASSLA-Stanza, followed by an additional analysis of model performance on specific labels. 


CLASSLA-Stanza gives consistent results across all supported languages and outperforms the Stanza pipeline on all supported NLP tasks, as illustrated in sections \ref{sec:clasvsstan} and \ref{sec:modtrapro}. However, overall low accuracies are still seen for infrequent labels and pairs of labels that are not so easily disambiguated. It remains to be seen whether larger and more diverse training datasets can contribute to improving model performance in these specific cases, or rather the move to contextual embeddings, i.e., transformer models. Additionally, when processing texts obtained from the internet, special care must be taken to use the combination of models that is best suited for the task, which is why we also described the special \textit{web} processing type implemented within CLASSLA-Stanza.


The release of a specialized pipeline for linguistic processing of South Slavic languages is an important new milestone in the development of digital resources and tools for this relatively under-resourced group of languages. However there is still much left to be achieved and improved upon. Full support for all annotation tasks, such as, for instance, semantic role labeling, which is currently only available for Slovenian, remains to be extended to other languages as well. As larger training datasets become available, more capable models can be trained for the currently supported languages. In addition, the aim is also to extend support to other members of the South Slavic language group, provided that training datasets of sufficient size are eventually produced for those languages as well. Finally, the performance of the CLASSLA-Stanza pipeline should also be compared to other recent state-of-the-art tools for automatic linguistic annotation, such as Trankit~\citep{trankit}, which was shown to outperform Stanza over a large number of languages and datasets.

\section{Acknowledgements}

The work described by this paper was made possible by the Development of Slovene in a Digital Environment project (Razvoj slovenščine v digitalnem okolju, project ID: C3340-20-278001), financed by the Ministry of Culture of the Republic of Slovenia and the European Regional Development Fund, the Language Resources and Technologies for Slovene research program (project ID: P6-0411), financed by the Slovenian Research Agency, the MEZZANINE project (Basic Research for the Development of Spoken Language Resources and Speech Technologies for the Slovenian Language, project ID: J7-4642), financed by the Slovenian Research Agency, and the CLARIN.SI research infrastructure. The work was also conducted as part of the MaCoCu action, which has received funding from the European Union’s Connecting Europe Facility 2014-2020 - CEF Telecom, under Grant Agreement No. INEA/CEF/ICT/A2020/2278341. The contents of this publication are the sole responsibility of its authors and do not necessarily reflect the opinion of the European Union.

\bibliographystyle{unsrtnat}
\bibliography{references}  






\end{document}